# RA-FinBERT: Rule-aware LoRA adaptation for low-resource financial sentiment classification

Fan Zhang† and Jiaming Li†,*

School of Informatics, Computing, and Cyber Systems, Northern Arizona University, Flagstaff, Arizona 86011, USA

† Co-first authors.

**Abstract:** Financial sentiment analysis converts unstructured financial news into quantitative signals that can support market analysis and decision-making. Existing work on resource-efficient financial NLP has largely focused on compressing or adapting pretrained language models, with less attention to combining contextual representations with lightweight rule-derived features. This study develops Rule-Aware FinBERT (RA-FinBERT), a parameter-efficient framework that integrates low-rank adaptation (LoRA) with three continuous VADER-derived sentiment proportions (positive, negative, and neutral) and a source-level metadata feature. The standardized four-dimensional feature vector is directly concatenated with the 768-dimensional final-layer FinBERT [CLS] representation and passed through a lightweight classification head. This design introduces only 1,024 additional trainable weights relative to a structurally matched text-only FinBERT model. RA-FinBERT was evaluated against text-only FinBERT and a lightweight DistilBERT baseline for three-class sentiment classification of financial-news titles and descriptions. On the held-out test set, RA-FinBERT achieved 69.89% accuracy and a macro F1 score of 0.634, compared with 63.44% and 0.526 for text-only FinBERT. Neutral-class recall increased from 18.18% to 45.45%. The framework supports both CPU and GPU execution, offering a lightweight and practical approach to financial sentiment classification under constrained computational resources. These findings indicate that rule-derived sentiment information and source metadata can provide complementary signals to contextual FinBERT representations and improve performance with minimal additional model complexity.

**Keywords:** financial sentiment analysis; FinBERT; LoRA; VADER; rule-aware feature fusion; weak supervision; resource-constrained NLP

## 1 Introduction

Financial sentiment analysis converts unstructured financial text into quantitative signals that can support market analysis and downstream forecasting. Early studies primarily relied on sentiment lexicons and conventional machine-learning methods. Tetlock used the Harvard Psychosociological Dictionary to analyse Wall Street Journal columns and showed that media pessimism was associated with subsequent downward pressure on stock prices, helping establish quantitative textual sentiment as an empirical tool in finance [1]. Loughran and McDonald later demonstrated that many words classified as negative by general-purpose dictionaries were not negative in financial contexts and developed a domain-specific financial sentiment dictionary that became widely used in subsequent research [2]. Supervised methods such as naive Bayes and support vector machines also showed that financial news and online discussions contained information relevant to market behaviour, although their reliance on hand-crafted features limited their ability to represent contextual and domain-dependent meaning [3].

The development of deep learning and pretrained language models substantially improved contextual representation in financial sentiment analysis [4]. BERT introduced bidirectional Transformer encoding and masked language modelling, leading to strong performance across a wide range of natural-language-understanding tasks [5]. Domain-adapted models such as FinBERT were subsequently developed to capture financial terminology and context [6]. However, domain adaptation does not guarantee superior performance across all datasets. Karanikola et al. reported that RoBERTa outperformed several conventional models, whereas FinBERT did not consistently outperform general-purpose language models [7]. Related studies combining BERT-based representations with recurrent, adversarial, or feature-augmented architectures have likewise shown that performance depends on the dataset, training strategy, and availability of complementary information [8–10]. These findings suggest that effective financial sentiment classification depends not only on pretrained representations but also on how those representations are adapted and combined with task-relevant information.

*Correspondence to: Jiaming Li, School of Informatics, Computing, and Cyber Systems, Northern Arizona University, Flagstaff, Arizona 86011, USA. E-mail: Jiaming.li@nau.edu

At the same time, growing interest in computationally efficient language modelling has led to methods that reduce the cost of adapting large pretrained models. Knowledge distillation, low-rank adaptation, quantization, few-shot prompting, and proxy tuning can substantially reduce the number of trainable parameters or the memory required for downstream adaptation [11–15]. LoRA is particularly attractive in small-data settings because it freezes most pretrained parameters while learning low-rank updates within selected linear transformations [16]. This allows domain-specific adaptation with a relatively small trainable parameter budget and supports efficient execution on both GPUs and commodity CPUs.

A complementary direction is to combine contextual language representations with lightweight structured information. Rule-based sentiment tools such as VADER are commonly used as standalone baselines or to derive categorical sentiment labels, yet their continuous intermediate outputs can also provide compact information about the composition of a sentiment assessment [17]. In particular, the positive, negative, and neutral proportions provide different views of the polarity distribution within a text. These features can be incorporated into a pretrained language model with minimal additional model complexity and may provide useful auxiliary signals when labelled data are limited. However, the value of directly integrating lightweight rule-derived features with parameter-efficient financial language models remains comparatively underexplored.

To address this gap, this study develops Rule-Aware FinBERT (RA-FinBERT), a parameter-efficient feature-fusion framework that combines a LoRA-adapted FinBERT representation with three VADER-derived sentiment proportions, namely positive, negative, and neutral, together with a source-level metadata feature. The standardized four-dimensional feature vector is directly concatenated with the final-layer FinBERT [CLS] representation and passed through a lightweight classification head for three-class sentiment prediction. The study makes three main contributions. First, it introduces a simple rule-aware fusion strategy that incorporates structured sentiment and source information while adding only 1,024 trainable weights relative to a structurally matched text-only FinBERT classifier. Second, it evaluates whether these auxiliary features provide complementary information beyond contextual FinBERT representations in a small-sample setting, using text-only FinBERT as the principal matched comparison and DistilBERT as a secondary lightweight baseline [18]. Third, it demonstrates a computationally efficient implementation based on LoRA that can be executed on either GPU or commodity CPU hardware, providing a practical framework for financial sentiment analysis when both labelled data and computational resources are limited.

## 2 Methods

### 2.1 Model overview and comparison design

RA-FinBERT combines a finance-domain text encoder, LoRA-based adaptation, and a lightweight set of auxiliary numerical inputs. For each news item, the title and description are concatenated and tokenized to a maximum length of 128 tokens. FinBERT contains 12 Transformer encoder layers with hidden size 768, and the final-layer contextualized hidden state at the [CLS] position is used as the text representation. LoRA modules are inserted into the Query and Value projections of all 12 layers. In parallel, the model receives three VADER-derived sentiment proportions, namely pos, neg, and neu, together with the source-level metadata feature

source_weight. The compound score is excluded from the final input set. The standardized four-dimensional numerical vector is directly concatenated with the 768-dimensional [CLS] representation to form a 772-dimensional classifier input. No separate numerical encoder is used.

The principal comparator is a structurally matched text-only FinBERT model. It uses the same pretrained checkpoint, final-layer [CLS] extraction, LoRA target layers, loss-weighting strategy, and two-layer task-specific classifier. Its classification head maps 768 dimensions to 256 and then to three classes, whereas the RA-FinBERT head maps 772 dimensions to 256 and then to three classes. This matched design makes the four auxiliary numerical inputs the principal architectural difference between the two models. The original FinBERT pooler and pretrained classification head are not used in the task-specific forward path. DistilBERT is retained as a secondary lightweight baseline for comparison.

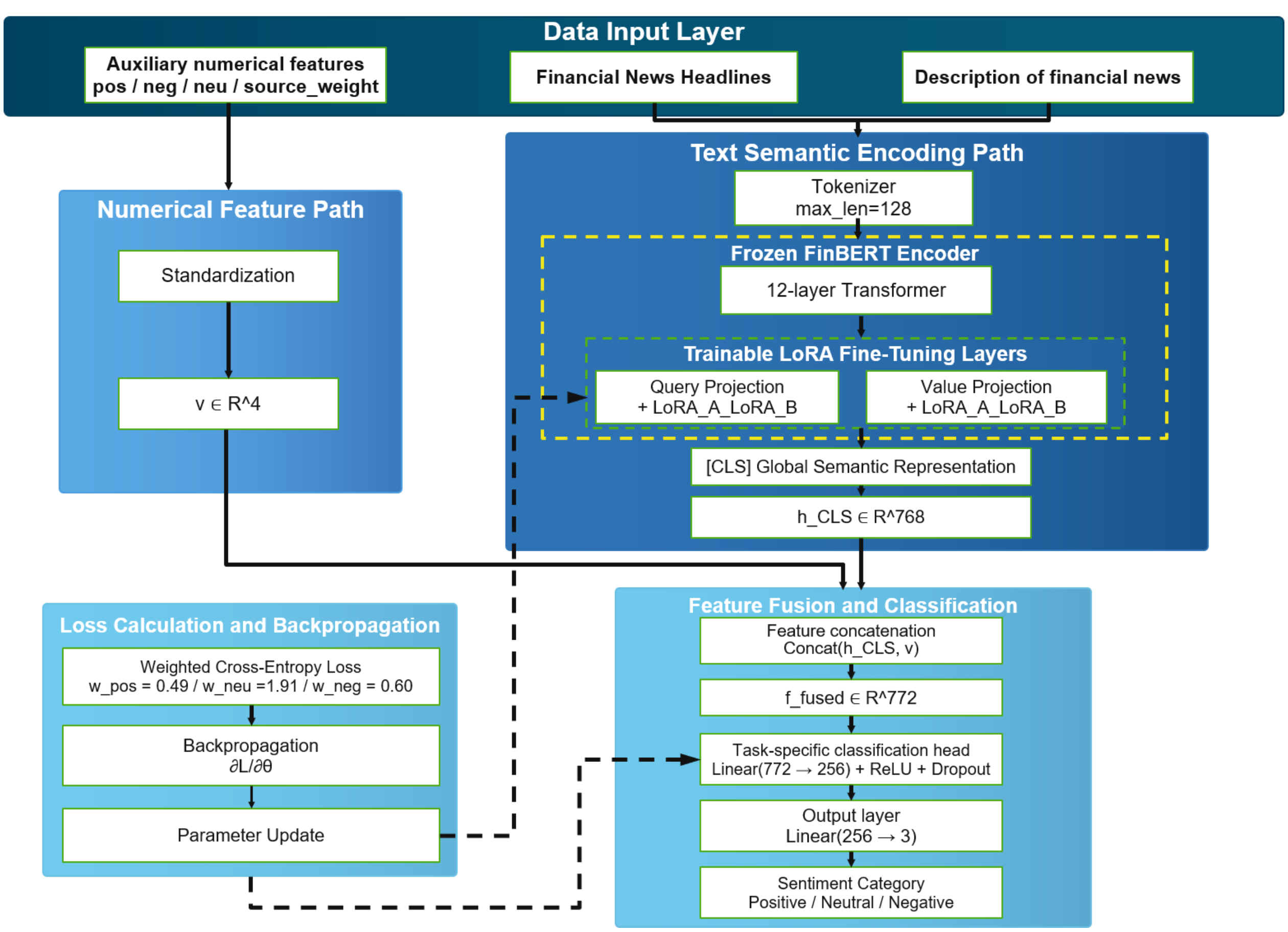

**Figure 1.** Architecture of RA-FinBERT. Financial-news titles and descriptions are encoded by FinBERT with LoRA applied to the Query and Value projections. The final-layer 768-dimensional [CLS] representation is directly concatenated with the standardized four-dimensional auxiliary vector comprising pos, neg, neu and source_weight, producing a 772-dimensional fused representation. The fused representation is passed through a 256-unit ReLU layer with dropout and a three-class output layer.

## 2.2 Text encoding and low-rank adaptation

Scaled dot-product attention supplies the basic Transformer operation [19]:

$$Attention(Q, K, V) = softmax\left(\frac{QK^T}{\sqrt{d_k}}\right)V \tag{1}$$

where $Q$, $K$ and $V$ are the Query, Key and Value matrices, and $d_k$ is the Key dimension. The FinBERT checkpoint used in this study was pretrained on financial corpora using a masked language-modelling objective [6]:

$$L_{MLM} = -\left(\frac{1}{M}\right)\sum_{i=1}^{M} \log p(x_i | x_{\text{masked}}) \tag{2}$$

where $M$ is the number of masked tokens and $x_{\text{masked}}$ is the masked input. During downstream adaptation, LoRA keeps the original weight matrix $W$ frozen and adds a low-rank update through matrices $A$ and $B$ [16]:

$$h = Wx + \left(\frac{\alpha}{r}\right) BAx \tag{3}$$

where $r$ denotes the LoRA rank and $\alpha$ is the scaling factor. LoRA was applied to the Query and Value projections in all 12 FinBERT layers with rank r = 16 and scaling factor α = 16. The pretrained Query and Value weight matrices remain frozen. Training jointly optimizes the LoRA A/B matrices, the corresponding Query/Value bias terms and the task-specific classification head. The remaining pretrained backbone parameters stay frozen.

### 2.3 Rule-derived features and direct fusion

VADER provides continuous positive, negative and neutral sentiment proportions in addition to its normalized compound score [17]. Let $S$ denote the sum of rule-adjusted token valences and a = 15 the VADER normalization constant:

$$S = \sum_{i=1}^{m} s_i, \quad compound = \frac{S}{\sqrt{S^2 + a}} \tag{4}$$

where $m$ is the number of valenced tokens and $s_i$ denotes the rule-adjusted valence of token i. For each concatenated news text x, the final-layer FinBERT hidden state at the [CLS] position provides the contextualized text representation:

$$h_{CLS} = FinBERT_{LoRA(x)_{[CLS]}} \in \mathbb{R}^{768} \tag{5}$$

The auxiliary numerical vector contains three VADER-derived sentiment proportions and one source-level metadata feature:

$$v = [pos, neg, neu, source_weight]^T \in \mathbb{R}^4 \tag{6}$$

StandardScaler was fitted only on the training partition and then applied to the validation and test partitions. For feature $j$, standardization is defined as:

$$v_j' = \frac{v_j - \mu_j}{\sigma_j} \tag{7}$$

where $\mu_j$ and $\sigma_j$ are estimated from the training partition. During training, the complete standardized numerical vector is zeroed with probability $p_{mask} = 0.1$ by multiplication with a Bernoulli variable:

$$\hat{v} = mv', \quad m \sim Bernoulli(0.9) \tag{8}$$

Whole-vector masking is used only as a training regularizer intended to discourage excessive dependence on the auxiliary inputs. No numerical masking is applied during validation or testing, so evaluation uses complete and deterministic numerical vectors. Accordingly, $\hat{v} = v'$ during validation and testing.

The final-layer [CLS] state and the numerical vector supplied to the model are concatenated directly:

$$f = [h_{CLS}; \hat{v}] \in \mathbb{R}^{772} \tag{9}$$

The fused representation is processed by a 256-unit ReLU layer with a dropout rate of 0.1, followed by a three-class output layer:

$$u = Dropout\big(ReLU(W_1 f + b_1)\big) \tag{10}$$

$$z = W_2 u + b_2 \in \mathbb{R}^3 \tag{11}$$

Softmax converts the logits to class probabilities:

$$p(y = c|x, \hat{v}) = \frac{\exp(z_c)}{\sum_{k=1}^{3} \exp(z_k)} \tag{12}$$

Direct concatenation changes the first classifier layer from 768 × 256 to 772 × 256. The four numerical inputs therefore add exactly 1,024 trainable weights relative to the matched text-only classifier.

### 2.4 Objective and label construction

Text-only FinBERT and RA-FinBERT use weighted cross-entropy. The class weights computed from the training set were 0.4919 for positive, 1.9120 for neutral and 0.5961 for negative. For $N$ examples, the objective is:

$$L = -\left(\frac{1}{N}\right) \sum_{i=1}^{N} w_{y_i} \log p(y_i|x_i, \hat{v}_i) \tag{13}$$

For the text-only FinBERT comparator, the same objective is used with the auxiliary numerical input $\hat{v}_i$ omitted. Let $c$ denote the VADER compound score. The categorical target sentiment_cat is generated by thresholding $c$:

$$y = positive\ (c \geq 0.05);\ neutral\ (|c| < 0.05);\ negative\ (c \leq -0.05) \tag{14}$$

Because compound directly determines the target labels, it was not included among the auxiliary model inputs. RA-FinBERT instead incorporates the continuous pos, neg and neu proportions together with source_weight as complementary features.

## 3 Experimental design

### 3.1 Hardware and software environment

The implementation was designed to operate under modest computational resources and was evaluated in both CPU-only and GPU-accelerated environments. CPU execution was performed on a consumer-grade laptop equipped with an Intel Core Ultra 5 225H processor and 32 GB of memory, without a discrete GPU. Under this setting, a complete 10-epoch training run required approximately 1-2 h. GPU-accelerated experiments were conducted in Google Colab using an NVIDIA T4 GPU, where the corresponding workflow completed in approximately 5-10 min. These runtimes are reported descriptively rather than as formal performance benchmarks. The implementation used a lightweight software stack based on PyTorch, Transformers, pandas, NumPy and scikit-learn, without distributed training or specialized large-scale acceleration frameworks.

### 3.2 Dataset and preprocessing

The public Kaggle dataset Financial News Sentiment vs Market 2020 was used. The selected CSV file contained financial-news titles, descriptions, publication metadata, VADER scores and sentiment categories. The analysed dataset contained 614 examples. Titles and descriptions were concatenated with a single intervening space. Stop-word filtering and stemming were not applied, preserving the input distribution expected by the pretrained tokenizer.

The dataset contained 295 positive examples (48.0%), 76 neutral examples (12.4%) and 243 negative examples (39.6%). The neutral class was strongly under-represented. The concatenated text had a mean length of 268.9 characters, a median of 275.5 and a maximum of 480. Given the relatively short news texts, a maximum sequence length of 128 tokens was used to retain the substantive content of most samples while limiting computational cost. Longer sequences were truncated by the tokenizer. By definition, VADER compound scores are bounded between −1 and 1, while the positive, negative and neutral proportions are bounded between 0 and 1. The source_weight metadata feature took values of 1.0 and 1.5.

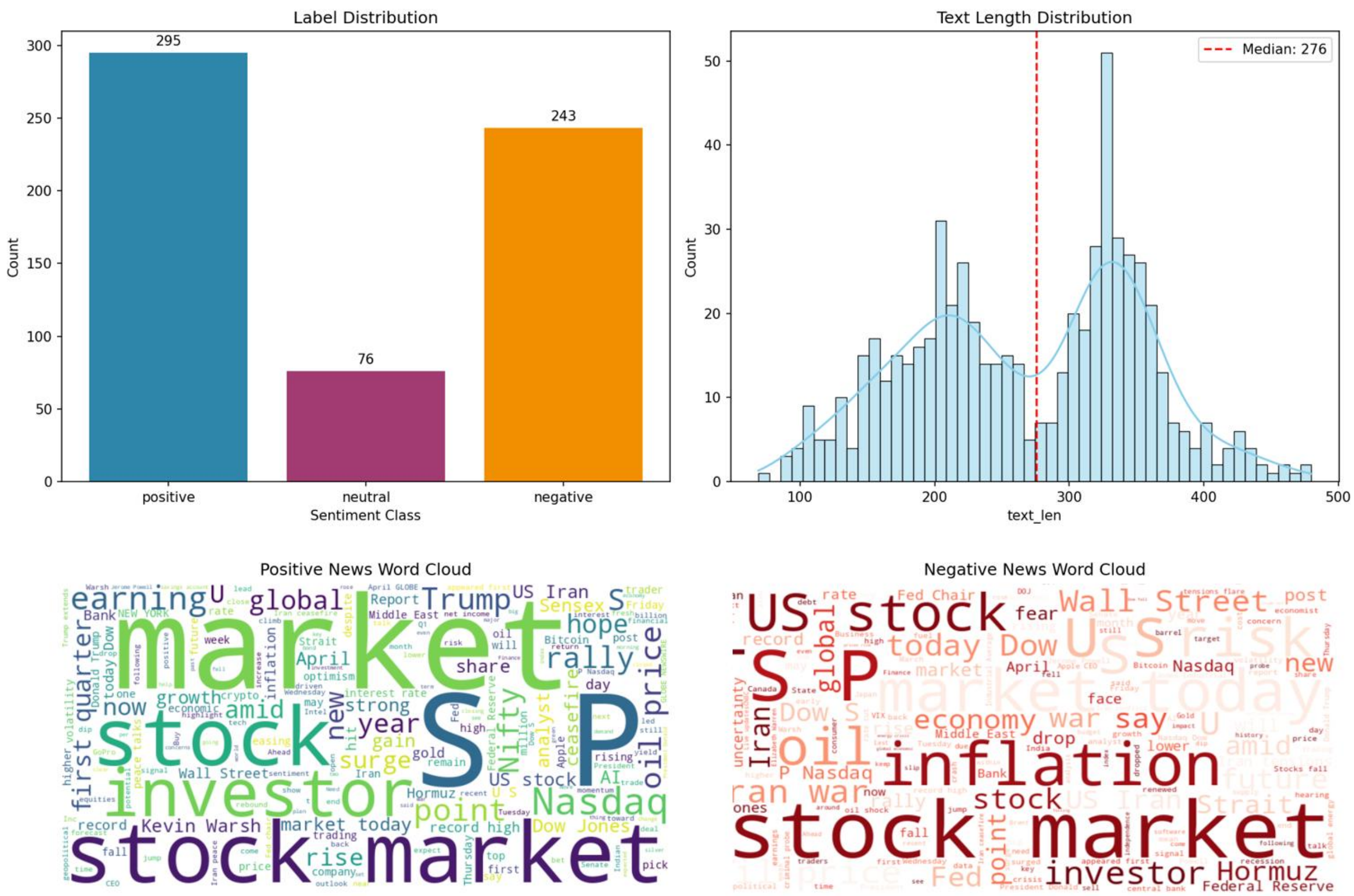

**Figure 2.** Dataset profile. The panels show class counts, text-length distribution, and word clouds for positive and negative news. Word-cloud size reflects term frequency and is descriptive rather than inferential.

A stratified split with random seed 42 produced 429 training examples, 92 validation examples and 93 test examples. StandardScaler was fitted to the four numerical columns in the training set, and the learned statistics were reused for validation and test data. Labels were mapped as positive = 0, neutral = 1 and negative = 2. The test supports were 45 positive, 11 neutral and 37 negative examples.

### 3.3 Models, training and evaluation

Shared hyperparameters were a maximum sequence length of 128, batch size 8, 10 training epochs, learning rate $\eta = 1 \times 10^{-4}$, LoRA rank $r = 16$, LoRA scaling $\alpha = 16$, early-stopping patience of 5 and random seed 42. AdamW used a weight decay of 0.01, and the gradient norm was clipped at 1.0. The best checkpoint for each model was selected according to validation accuracy. All three models used LoRA, so table and figure labels omit the redundant "+ LoRA" suffix.

DistilBERT used WeightedRandomSampler with inverse-frequency sample weights and ordinary cross-entropy. Text-only FinBERT and RA-FinBERT used shuffled batches and the same class-weighted cross-entropy. Both strategies were used to mitigate the strong under-representation of the neutral class, with sampling increasing minority-class exposure for DistilBERT and loss weighting increasing the penalty for minority-class errors in the two FinBERT models. Both FinBERT models used the final-layer contextualized [CLS] vector and a custom 256-unit classifier. RA-FinBERT differed by concatenating the four standardized numerical features before that classifier. Backpropagation jointly updated LoRA A/B, the corresponding Q/V biases and each task-specific head.

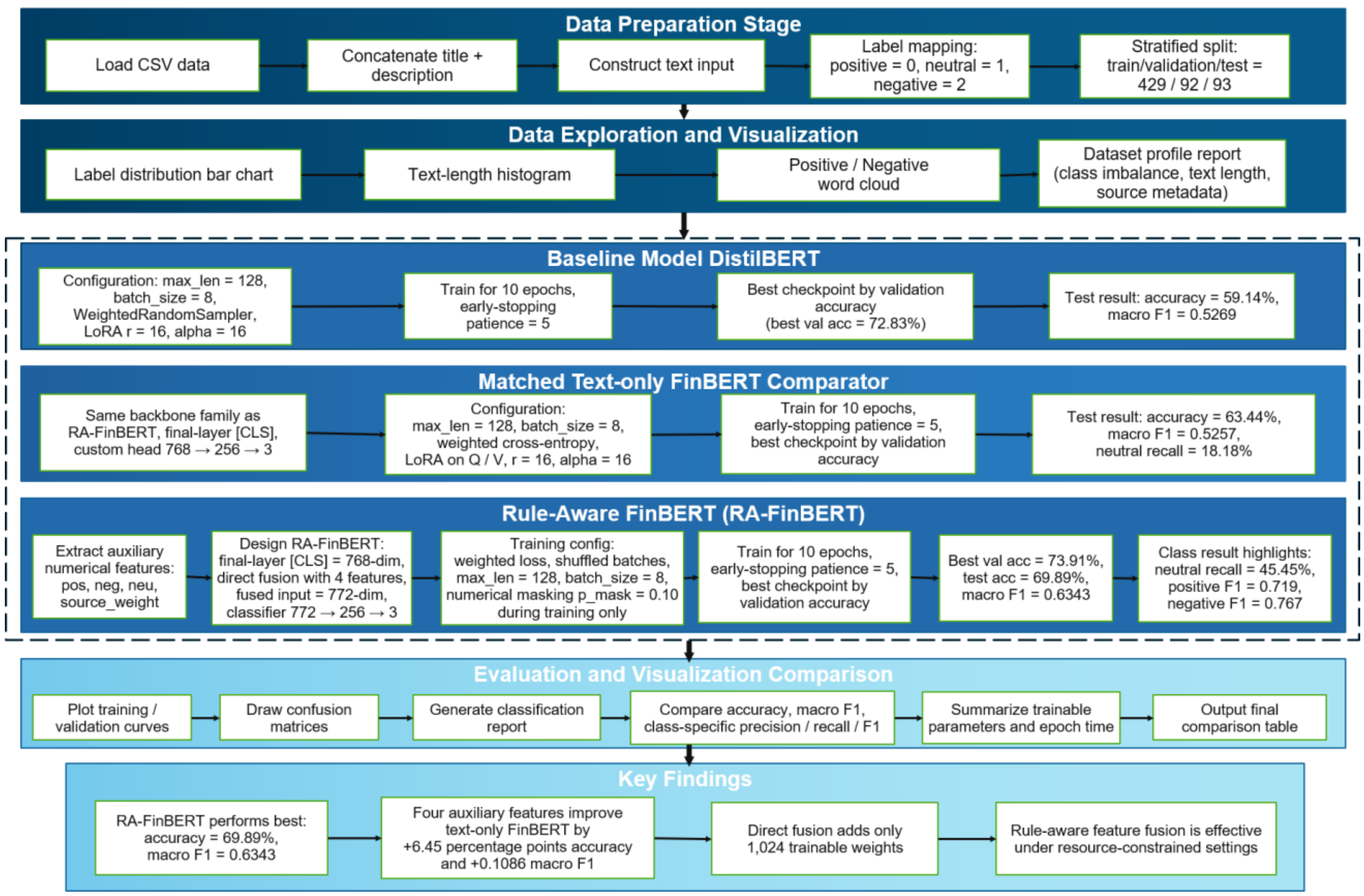

**Figure 3.** Experimental workflow. The three models share the data split, maximum sequence length, ten-epoch training window, learning rate, LoRA rank and scaling, and random seed. Their backbone, tokenizer, imbalance strategy and auxiliary-feature input differ.

The principal comparison is text-only FinBERT versus RA-FinBERT because their shared backbone and matched heads more directly isolate the contribution of the four auxiliary inputs. DistilBERT is a secondary lightweight baseline and differs in backbone and classifier implementation. Evaluation reports test accuracy, macro F1, class-specific precision, recall and F1, and confusion matrices. Validation macro F1 was calculated descriptively but was not used for checkpoint selection.

## 4 Results and discussion

### 4.1 Optimisation behaviour

Across the ten epochs, RA-FinBERT training loss decreased steadily. Validation accuracy reached 0.7391 at epochs 5 to 7 and ended at 0.7283. Validation loss declined through approximately epoch 7 and then rose moderately, consistent with mild late-epoch overfitting. Text-only FinBERT reached its highest validation accuracy at epoch 9 (0.7500) and ended at 0.7283, while its validation loss rose after the middle epochs as training loss continued to decline. DistilBERT peaked at 0.7283 validation accuracy at epoch 6 and ended at 0.6522; its validation loss also increased late in training.

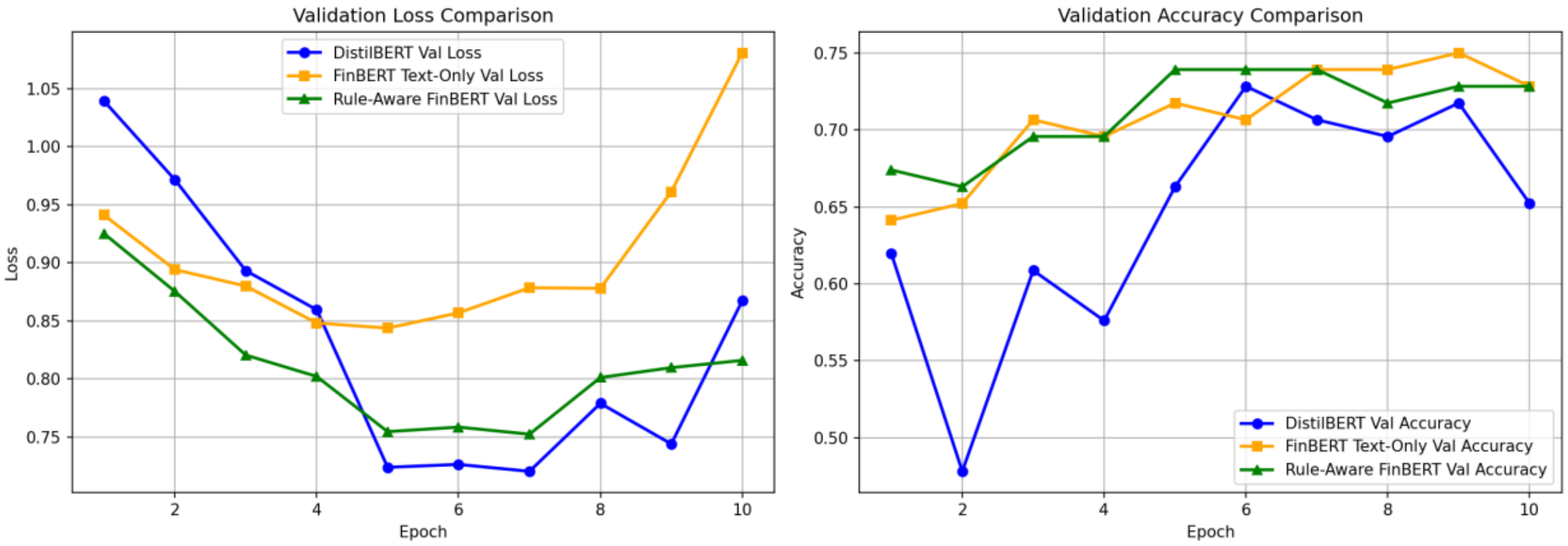

**Figure 4.** Validation trajectories across ten epochs. The left panel compares validation loss and the right panel compares validation accuracy for one stratified split and random seed.

## 4.2 Comparative test performance

On the held-out test set, RA-FinBERT achieved 0.6989 accuracy and a macro F1 score of 0.6343. The matched text-only FinBERT model achieved 0.6344 accuracy and 0.5257 macro F1. The observed differences were 6.45 percentage points in accuracy and 0.1086 in macro F1. DistilBERT, retained as a secondary lightweight baseline, achieved 0.5914 accuracy and 0.5269 macro F1. By accuracy, RA-FinBERT ranked first, followed by text-only FinBERT and DistilBERT. RA-FinBERT also had the highest macro F1, whereas the macro F1 values for the two baselines were nearly identical.

**Table 1.** Test performance. All models were fine-tuned with LoRA; model names omit the suffix for clarity. Class-specific values are computed from the final test confusion matrices.

| *Model* | *Accuracy* | *Macro F1* | *Positive F1* | *Neutral F1* | *Negative F1* | *Neutral recall* |
|---|---|---|---|---|---|---|
| *DistilBERT* | 0.5914 | 0.5269 | 0.6265 | 0.2500 | 0.7042 | 0.3636 |
| *Text-only FinBERT* | 0.6344 | 0.5257 | 0.6800 | 0.2000 | 0.6970 | 0.1818 |
| *RA-FinBERT* | 0.6989 | 0.6343 | 0.7191 | 0.4167 | 0.7671 | 0.4545 |

RA-FinBERT correctly classified 32 of 45 positive, 5 of 11 neutral and 28 of 37 negative test examples, for 65 correct predictions. Text-only FinBERT correctly classified 34, 2 and 23 examples, respectively, for 59 correct predictions. RA-FinBERT therefore produced six additional correct predictions overall. Neutral recall increased from 18.18% to 45.45%, corresponding to three additional correctly classified neutral items, and negative recall increased from 62.16% to 75.68%. Positive recall decreased from 75.56% to 71.11%, so the improvement was not uniform across classes.

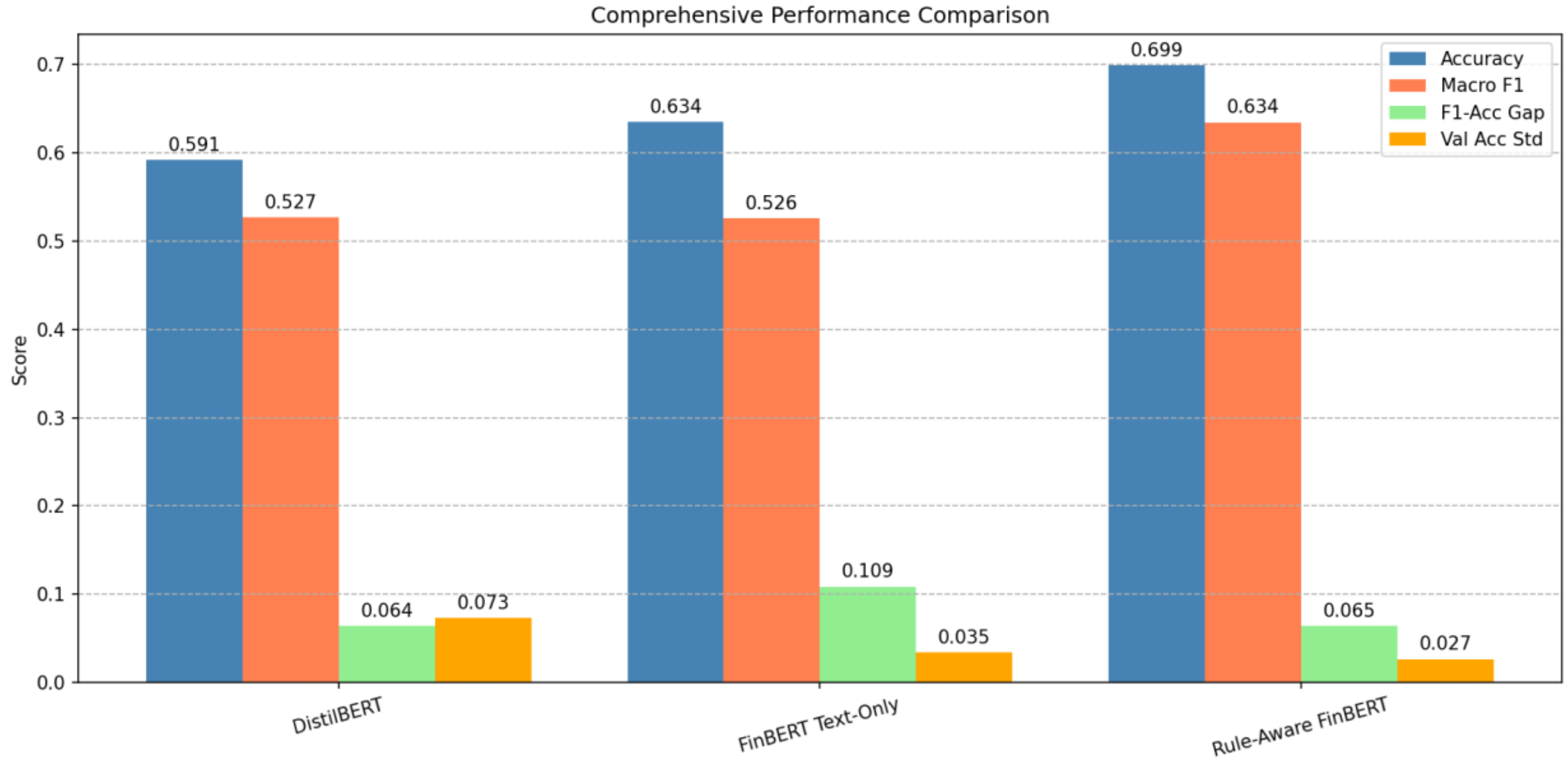

**Figure 5.** Model-level descriptive metrics. Test accuracy and macro F1 are the primary performance summaries. The F1-accuracy gap and the standard deviation of validation accuracy across training epochs are shown as auxiliary diagnostics; smaller values indicate closer agreement between the two test metrics and lower epoch-to-epoch variation, respectively, rather than superior predictive performance.

Relative to text-only FinBERT, the RA-FinBERT confusion matrix contained fewer neutral-to-positive errors (4 versus 8) and fewer negative-to-positive errors (8 versus 13). Positive-to-neutral errors increased from 6 to 7, positive-to-negative errors increased from 5 to 6, and neutral-to-negative errors increased from 1 to 2; negative-to-neutral errors remained at 1. Differences from DistilBERT cannot be attributed solely to the numerical features because the backbone and classifier also differ.

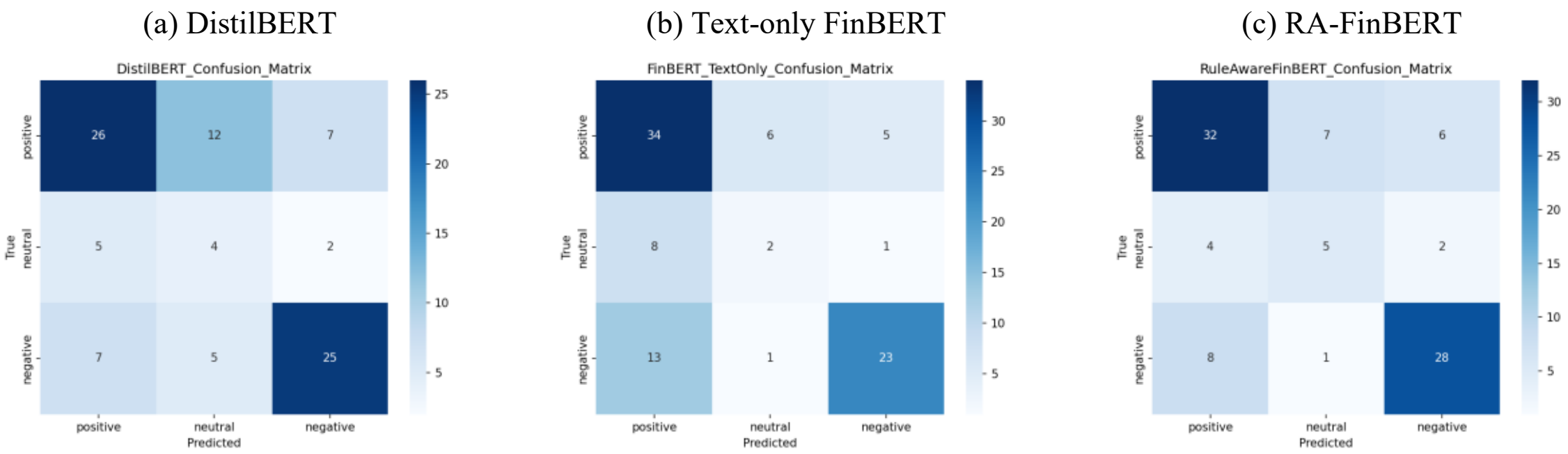

**Figure 6.** Test-set confusion matrices for DistilBERT, text-only FinBERT and RA-FinBERT. Rows are true classes and columns are predicted classes; class order is positive, neutral and negative.

### 4.3 Feature fusion and parameter efficiency

The auxiliary vector comprised the pos, neg and neu proportions together with source_weight. The three proportions summarize the composition of VADER's polarity assessment, while source_weight contributes source-level metadata. Direct concatenation allows the classifier to integrate these lightweight signals with the contextualized [CLS] representation without a separate numerical encoder. This design introduces rule-derived sentiment and metadata information with minimal additional model complexity.

The F1-accuracy gaps were approximately 0.064 for DistilBERT, 0.109 for text-only FinBERT and 0.065 for RA-FinBERT. Validation-accuracy standard deviations across epochs were approximately 0.073, 0.035 and 0.027, respectively. These secondary diagnostics describe agreement between aggregate metrics and epoch-to-epoch variability and are interpreted separately from the primary test-performance measures.

Parameter counts confirm the low overhead of direct fusion. Text-only FinBERT contained 110,272,006 parameters, of which 805,891 were trainable. RA-FinBERT contained 110,273,030 parameters, of which 806,915 were trainable. The difference is exactly 1,024 trainable weights, arising from four additional inputs to the 256-unit first classifier layer. The trainable fraction remained approximately 0.73% for both models. DistilBERT contained 67,250,691 parameters and 897,027 trainable parameters (1.33%).

**Table 2.** Model size and optimization strategy. Counts include task-specific classification heads; all models used LoRA. Epoch times are descriptive averages from the evaluated Colab T4 run.

| *Model* | *Total parameters* | *Trainable* | *Trainable (%)* | *Imbalance strategy* | *Avg epoch time (s)* |
|---|---|---|---|---|---|
| *DistilBERT* | 67,250,691 | 897,027 | 1.33% | Balanced sampling | 3.68 |
| *Text-only FinBERT* | 110,272,006 | 805,891 | 0.73% | Weighted loss | 7.27 |
| *RA-FinBERT* | 110,273,030 | 806,915 | 0.73% | Weighted loss | 7.31 |

## 4.4 Limitations and future work

Several limitations constrain interpretation. The dataset contained only 614 examples, and the test set contained 93 examples, including just 11 neutral items. One additional correct neutral prediction changes neutral recall by approximately 9.1 percentage points. Results from one primary split and one seed do not quantify uncertainty across samples or initializations; repeated stratified runs, cross-validation or bootstrap confidence intervals are needed before claims of consistent superiority can be made.

The target labels are VADER-derived weak labels rather than independent human annotations. Compound deterministically reproduced all 614 labels and was excluded from the final model, removing the direct target-construction variable. However, pos, neg and neu remain outputs of the same VADER system. Their relationship to the weak labels is therefore not eliminated, and the observed results cannot establish transfer to human financial-sentiment judgments.

Whole-vector masking was applied only during training and was disabled for validation and testing. This design provides deterministic evaluation inputs, but the 0.10 masking probability is only a regularization choice and does not demonstrate robustness or remove all dependence between rule-derived inputs and targets. The auxiliary F1-accuracy and validation-variability diagnostics are likewise descriptive and do not replace repeated-run uncertainty estimates.

The comparison also used different imbalance handling across backbones: DistilBERT used balanced sampling, whereas both FinBERT models used weighted loss. The matched FinBERT comparison remains the appropriate basis for interpreting the auxiliary features, while differences involving DistilBERT reflect multiple architectural and optimization factors. Future work should evaluate larger independently annotated datasets across different time periods, market conditions and news sources, repeat training across multiple random seeds, perform ablation analyses of the individual auxiliary features and numerical masking strategy, and compare imbalance-handling approaches under matched experimental settings. Further studies could also examine alternative LoRA ranks and insertion layers, compare LoRA with other parameter-efficient adaptation methods such as adapters or prefix tuning [20], and investigate alternative fusion architectures, including lightweight numerical encoders, attention-based fusion and gating mechanisms, particularly as richer and potentially higher-dimensional structured information becomes available.

# 5 Conclusion

This study developed RA-FinBERT as a lightweight rule-aware framework for combining domain-adapted language representations with structured auxiliary knowledge. The approach couples LoRA-based adaptation of FinBERT with direct fusion of three VADER-derived sentiment proportions and a source-level metadata feature.

Rather than introducing a separate numerical encoder or extensively fine-tuning the pretrained backbone, the proposed design preserves the contextual representation learned by FinBERT while allowing compact rule-derived signals to contribute directly to downstream classification. The resulting architecture adds only 1,024 trainable weights relative to a structurally matched text-only FinBERT classifier, while retaining a trainable fraction of approximately 0.73% of the full model.

On the held-out test set, RA-FinBERT achieved 69.89% accuracy and a macro F1 score of 0.6343, compared with 63.44% and 0.5257 for the matched text-only FinBERT model. The improvement was accompanied by higher neutral and negative recall, while positive recall decreased slightly, indicating that the gains were driven mainly by improved neutral and negative classification, particularly for the minority neutral class. Together with the small increase in trainable parameters and compatibility with both CPU-only and GPU-accelerated execution, these findings show that lightweight auxiliary information can complement contextual language representations without requiring substantial additional model complexity.

More broadly, the contribution of this work is a reusable methodological template for integrating pretrained domain models with existing rule systems, expert-designed indicators or other low-dimensional structured signals. Many application domains already contain useful prior knowledge in the form of scoring rules, heuristics, metadata, domain indices or expert-defined features. Rather than discarding such knowledge when adopting pretrained language models, a rule-aware framework can incorporate it as complementary information while preserving the representational capacity of the language model and the computational advantages of parameter-efficient adaptation. RA-FinBERT provides an initial demonstration of this principle in financial sentiment analysis and suggests a practical direction for resource-constrained domain NLP in settings where labelled data, computational resources or both are limited. Future work should evaluate this framework on larger independently annotated datasets, perform feature-level ablation analyses and repeated runs, and examine whether similar rule-aware fusion strategies transfer to other domain-specific language tasks.

## Conflict of Interest

The authors declare no conflicts of interest.

## Data Availability Statement

The underlying dataset used in this study is publicly available through Kaggle. The processed analytic data and the code used for data preprocessing, model training and evaluation are available from the corresponding author upon request.

## References

[1] Tetlock, P. C. Giving content to investor sentiment: The role of media in the stock market. J. Finance 62, 1139-1168 (2007). https://doi.org/10.1111/j.1540-6261.2007.01232.x

[2] Loughran, T. & McDonald, B. When is a liability not a liability? Textual analysis, dictionaries, and 10-Ks. J. Finance 66, 35-65 (2011). https://doi.org/10.1111/j.1540-6261.2010.01625.x

[3] Antweiler, W. & Frank, M. Z. Is all that talk just noise? The information content of internet stock message boards. J. Finance 59, 1259-1294 (2004). https://doi.org/10.1111/j.1540-6261.2004.00662.x

[4] Sohangir, S., Wang, D., Pomeranets, A. & Khoshgoftaar, T. M. Big data: Deep learning for financial sentiment analysis. J. Big Data 5, 3 (2018). https://doi.org/10.1186/s40537-017-0111-6

[5] Devlin, J., Chang, M.-W., Lee, K. & Toutanova, K. BERT: Pre-training of deep bidirectional transformers for language understanding. In Proc. 2019 Conf. North American Chapter of the Association for Computational Linguistics: Human Language Technologies 4171-4186 (Association for Computational Linguistics, 2019). https://doi.org/10.18653/v1/N19-1423

[6] Araci, D. FinBERT: Financial sentiment analysis with pre-trained language models. Preprint at https://doi.org/10.48550/arXiv.1908.10063 (2019).

[7] Karanikola, A., Davrazos, G., Liapis, C. M. & Kotsiantis, S. Financial sentiment analysis: Classic methods vs. deep learning models. Intell. Decis. Technol. 17, 893-915 (2023). https://doi.org/10.3233/IDT-230478

[8] Zhu, H., Lu, X. F. & Xue, L. Emotional analysis model of financial text based on the BERT. J. Shanghai Univ. (Nat. Sci. Ed.) 29, 118–128 (2023). https://doi.org/10.12066/j.issn.1007-2861.2308

[9] Duan, W. C. & Xue, T. FinBERT-RCNN-ATTACK: Emotional analysis model of financial text. Comput. Technol. Dev. 34, 157–162 (2024).

[10] Xu, X. C. & Tian, K. A novel financial text sentiment analysis-based approach for stock index prediction. J. Quant. Tech. Econ. 38, 124-145 (2021). https://doi.org/10.13653/j.cnki.jqte.2021.12.009

[11] Konstantinidis, T., Iacovides, G., Xu, M., Constantinides, T. G. & Mandic, D. P. FinLlama: Financial sentiment classification for algorithmic trading applications. Preprint at https://doi.org/10.48550/arXiv.2403.12285 (2024).

[12] Pontes, E. L., González-Gallardo, C.-E., Benjannet, M., Qu, C. & Doucet, A. L3iTC at the FinLLM Challenge Task: Quantization for Financial Text Classification & Summarization. In Proc. Eighth Financial Technology and Natural Language Processing and the 1st Agent AI for Scenario Planning 141-145 (2024). https://aclanthology.org/2024.finnlp-2.14

[13] Huang, Y., Ma, T., Yang, K. & Zhang, Z. FinSent-DistillQ: A distilled large language model with chain-of-thought fine-tuning for financial sentiment analysis. J. Intell. Inf. Syst. 64, 735-771 (2026). https://doi.org/10.1007/s10844-025-01020-9

[14] Wang, Y., Wang, Y., Liu, Y., Bao, R., Harimoto, K. & Sun, X. Proxy tuning for financial sentiment analysis: Overcoming data scarcity and computational barriers. In Proc. Joint Workshop of the 9th Financial Technology and Natural Language Processing, the 6th Financial Narrative Processing, and the 1st Workshop on Large Language Models for Finance and Legal 169-174 (Association for Computational Linguistics, 2025). https://aclanthology.org/2025.finnlp-1.16

[15] Todd, A., Bowden, J. & Moshfeghi, Y. Text-based sentiment analysis in finance: Synthesising the existing literature and exploring future directions. Intell. Syst. Account. Finance Manag. 31, e1549 (2024). https://doi.org/10.1002/isaf.1549

[16] Hu, E. J. et al. LoRA: Low-rank adaptation of large language models. In International Conference on Learning Representations (2022). https://openreview.net/forum?id=nZeVKeeFYf9

[17] Hutto, C. J. & Gilbert, E. VADER: A parsimonious rule-based model for sentiment analysis of social media text. Proc. Int. AAAI Conf. Web Soc. Media 8, 216-225 (2014). https://doi.org/10.1609/icwsm.v8i1.14550

[18] Sanh, V., Debut, L., Chaumond, J. & Wolf, T. DistilBERT, a distilled version of BERT: Smaller, faster, cheaper and lighter. Preprint at https://doi.org/10.48550/arXiv.1910.01108 (2019).

[19] Vaswani, A. et al. Attention is all you need. Adv. Neural Inf. Process. Syst. 30, 5998-6008 (2017).

[20] Wu, C., Zhao, Y., Liu, X., Si, N., Zhang, L. & Fan, H. Fine tuning methods for large language models: A survey. Journal of Chinese Information Processing 39, 1–26 (2025).